\documentclass{article} % For LaTeX2e
\usepackage{nips14submit_e,times}
\usepackage{hyperref}
\usepackage{url}
\usepackage{multirow}

\usepackage{framed}
\usepackage{algorithmic}
\usepackage{graphicx}
\usepackage{amssymb}
\usepackage{amsmath}
\usepackage{amsthm}
\usepackage{caption}
\usepackage{subcaption}
\usepackage{listings}
\usepackage{verbatimbox}
\usepackage[normalsections,normalmargins,normalindent,normalleading,normaltitle,normalbib]{savetrees}
\usepackage[titletoc,title]{appendix}

\usepackage[compact]{titlesec}
\titlespacing{\section}{0pt}{0.5ex}{0.3ex}
\titlespacing{\subsection}{0pt}{0.2ex}{0ex}
\titlespacing{\subsubsection}{0pt}{0.1ex}{0ex}

\usepackage{enumitem}
\setlist[itemize]{leftmargin=*}

\makeatletter
\ifcase \@ptsize \relax% 10pt
  \newcommand{\miniscule}{\@setfontsize\miniscule{4}{5}}% \tiny: 5/6
\or% 11pt
  \newcommand{\miniscule}{\@setfontsize\miniscule{5}{6}}% \tiny: 6/7
\or% 12pt
  \newcommand{\miniscule}{\@setfontsize\miniscule{5}{6}}% \tiny: 6/7
\fi
\makeatother

\newcommand {\aplt} {\ {\raise-.5ex\hbox{$\buildrel<\over\sim$}}\ }

\newcommand{\eqn}[1]{Eqn.~\ref{eqn:#1}}
\newcommand{\fig}[1]{Fig.~\ref{fig:#1}}
\newcommand{\tab}[1]{Table~\ref{tab:#1}}
\newcommand{\secc}[1]{Section~\ref{sec:#1}}
\def\etal{{\textit{et~al.~}}}

\usepackage{footnote}
\makesavenoteenv{tabular}
\makesavenoteenv{table}

\title{End-To-End Memory Networks}

\author{
Sainbayar Sukhbaatar\\
Dept. of Computer Science\\
Courant Institute, New York University\\
\texttt{sainbar@cs.nyu.edu}
\And
Arthur Szlam \:\:\:\:\:\: Jason Weston \:\:\:\:\:\: Rob Fergus \\
Facebook AI Research \\
New York\\
\texttt{\{aszlam,jase,robfergus\}@fb.com}
}

\nipsfinalcopy % Uncomment for camera-ready version

\begin{document}

\maketitle

\vspace{-5mm}
\begin{abstract}
We introduce a neural network with a recurrent attention model over a possibly
large external memory.
The architecture is a form of Memory Network
\cite{Weston14} but unlike the model in that work, it is trained end-to-end,
and hence requires significantly less supervision during training, making it more generally
applicable in realistic settings.
It can also be seen as an extension of RNNsearch \cite{BahdanauCB14} to the case where
multiple computational steps (hops) are performed per output symbol.
The flexibility of the model allows us to apply it to tasks as diverse
as (synthetic) question answering \cite{Weston15} and to language modeling.
For the former our approach is competitive with Memory Networks, but with less supervision.
For the latter, on the Penn TreeBank and Text8 datasets our approach demonstrates
  comparable performance to RNNs and LSTMs.
In both cases we show that the key concept of
multiple computational hops yields improved results.
\end{abstract}

\section{Introduction}
Two grand challenges in artificial intelligence research have been to build
models that can make multiple computational steps in
the service of answering a question or completing a task, and models that
can describe long term dependencies in sequential data.

Recently there has been a resurgence in models of computation using explicit
storage and a notion of attention
\cite{Weston14, Graves14, BahdanauCB14}; manipulating such a storage offers an approach to both of these challenges.
In \cite{Weston14, Graves14, BahdanauCB14}, the storage is endowed with a
continuous representation; reads from and writes to the storage, as well as other processing steps, are modeled by the actions of neural networks.

In this work, we present a novel recurrent neural network (RNN)
 architecture where the recurrence reads from a possibly large external memory
multiple times before outputting a symbol.
Our model can be considered a continuous form of the Memory Network implemented in
\cite{Weston14}.   The model in that work was not easy to train via backpropagation, and required supervision at each layer of the network.  The continuity of the model we present here
means that it can be trained end-to-end from input-output pairs, and so
 is applicable to more tasks,
i.e. tasks where such supervision is not available, such as in language modeling or
 realistically supervised question answering tasks.  Our model can also be seen as
 a version of RNNsearch \cite{BahdanauCB14} with multiple computational steps  (which we term ``hops'')  per output symbol.
We will show experimentally that the  multiple hops over  the long-term memory are crucial to
good performance of our model on these tasks, and that training the memory representation can be integrated in a scalable manner into our
end-to-end neural network model.
\section{Approach}
\label{sec:approach}

Our model takes a discrete set of inputs $x_1, ..., x_n$ that are to be
stored in the memory, a query $q$, and outputs an answer $a$.
Each of the $x_i$, $q$, and $a$ contains symbols coming from a dictionary with $V$ words.
The model writes all $x$  to the memory up to a fixed buffer size, and then finds a continuous
representation for the $x$ and $q$.  The continuous representation is then processed via multiple hops to output $a$.  This allows
backpropagation of the error signal through multiple memory
accesses back to the input during training.

\subsection{Single Layer}
We start by describing our model in the single layer case, which implements a single memory hop operation. We then show it can be stacked to give multiple hops in memory.

\noindent {\bf Input memory  representation:} Suppose we are given an input
set $x_1 ,.., x_i$ to be stored in memory.
The entire set  of $\{x_i\}$
are converted into memory vectors $\{m_i\}$ of dimension $d$ computed by embedding each $x_i$ in a continuous space, in the simplest case,
using an embedding matrix $A$ (of size $d \times V$).
The query $q$
is also embedded (again, in the simplest case via another embedding matrix $B$ with the same dimensions
as $A$) to obtain an internal state $u$.  In the embedding space, we compute the
match between $u$ and each memory $m_i$ by taking the
inner product followed by a softmax:
\begin{equation} \label{eq:p}
p_i = \text{Softmax}(u^T m_i).
\end{equation}
where $\text{Softmax}(z_i)=e^{z_i}/\sum_j e^{z_j}$. Defined in this way $p$ is a probability vector over the inputs.

\noindent {\bf Output memory representation:} Each $x_i$ has a
corresponding output vector $c_i$ (given in the simplest case by another embedding matrix
$C$).
 The response vector from the memory $o$ is then a sum over the transformed inputs $c_i$, weighted by the probability vector from the input:
\begin{equation}
o = \sum_i p_i c_i.
\end{equation}
Because the function from input to output is
smooth, we can easily compute gradients and back-propagate through
it.
Other recently proposed forms of memory or attention take this
approach, notably Bahdanau \etal \cite{BahdanauCB14} and Graves \etal \cite{Graves14}, see also~\cite{GregorDGW15}.

\noindent {\bf Generating the final prediction:}
In the single layer case, the sum of the output vector $o$ and the input embedding $u$ is then passed through a final weight matrix $W$ (of size $V \times d$) and a softmax to produce the predicted label:
\begin{equation}
\hat{a} = \text{Softmax}(W (o + u))
\end{equation}

\begin{figure}[h!]
\vspace{-3mm}
\begin{center}
\includegraphics[width=1\linewidth]{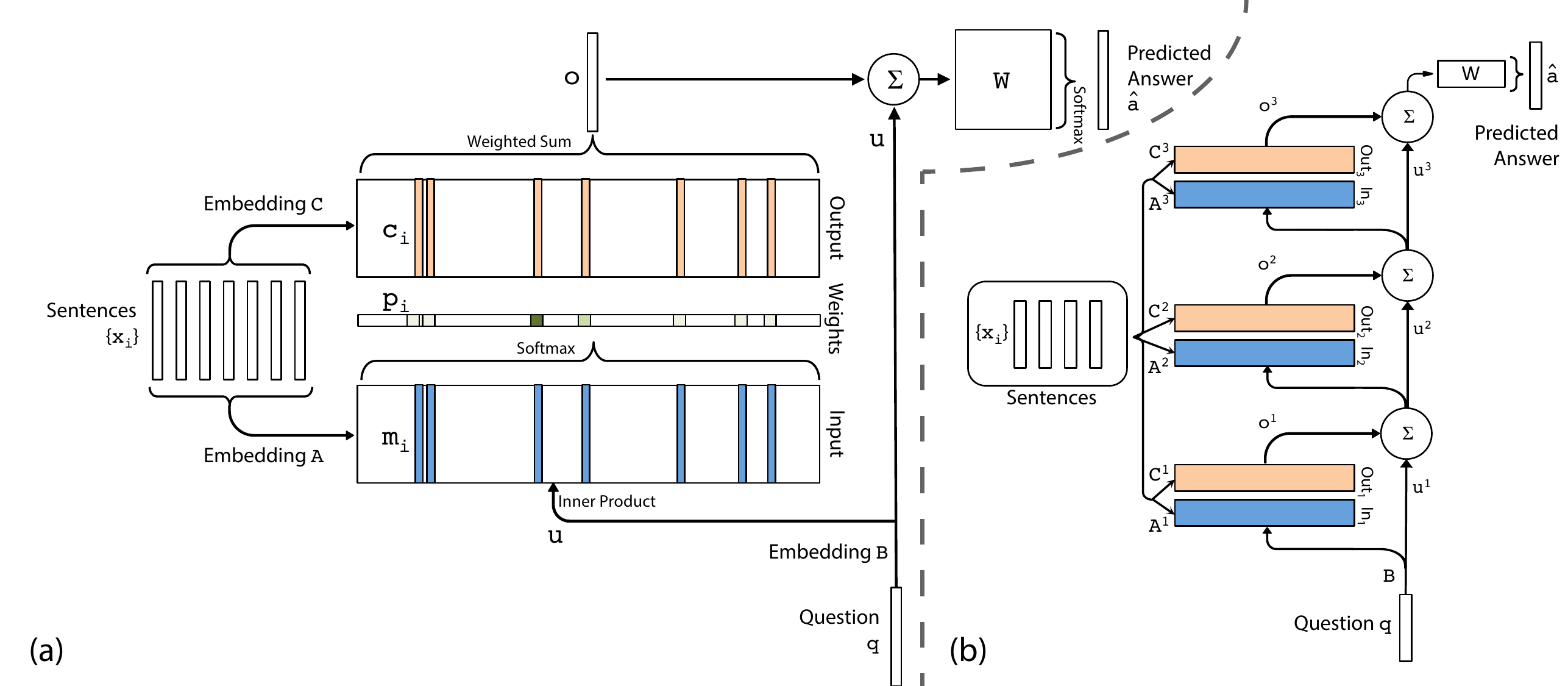}
\end{center}
\vspace*{-0.3cm}
\caption{(a): A single layer version of our model. (b): A three layer version of our model. In practice, we can constrain several of the embedding matrices to be the same (see \secc{multi}).}
\label{fig:single}
\vspace{-3mm}
\end{figure}

The overall model is shown in \fig{single}(a). During training, all three
embedding matrices $A$, $B$ and $C$, as well as $W$ are jointly
learned by minimizing a standard cross-entropy loss between $\hat{a}$
and the true label $a$. Training is performed using stochastic
gradient descent (see \secc{training} for more details).

\subsection{Multiple Layers}
\label{sec:multi}
We now extend our model to handle $K$ hop
operations. The memory layers are stacked in the following way:
\begin{itemize}
\item The input to layers above the first is the sum of the output $o^k$ and the input $u^{k}$ from layer $k$ (different ways to combine $o^k$ and $u^{k}$ are proposed later):
\begin{equation}\label{eqn:recurrent} u^{k+1} = u^k + o^k.\end{equation}
\item Each layer has its own embedding matrices $A^k,C^k$, used to
  embed the inputs $\{x_i\}$. However, as discussed below,
  they are constrained to ease training and reduce the number of parameters.
\item At the top of the network, the input to $W$ also combines the input and the output of the top memory layer: $\hat{a} = \text{Softmax}(W u^{K+1}) = \text{Softmax}(W(o^K + u^K))$.
\end{itemize}

We explore two types of weight tying within the model:
\begin{enumerate}
\item {\bf Adjacent:} the output embedding for one layer is the input
  embedding for the one above, i.e.~$A^{k+1}=C^k$. We also constrain (a) the answer prediction matrix to be
the same as the final output embedding, i.e~$W^T=C^K$, and (b) the
question embedding to match the input embedding of the first layer, i.e.~$B=A^1$.
\item {\bf Layer-wise (RNN-like):} the input and output embeddings are the same
  across different layers, i.e.~$A^1=A^2=...=A^K$ and $C^1=C^2=...=C^K$. We have found it useful to add a linear mapping $H$ to the update of $u$ between
hops; that is, $u^{k+1} = Hu^k + o^k$. This mapping is learnt along with the rest of the parameters and used throughout our experiments for layer-wise weight tying.
\end{enumerate}
A three-layer version of our memory model is shown in
\fig{single}(b). Overall, it is similar to the Memory Network model in \cite{Weston14}, except that the hard max operations within each layer have
been replaced with a continuous weighting from the
softmax.

Note that if we use the layer-wise weight tying scheme, our model can
be cast as a traditional RNN where we divide the outputs of the RNN
into {\it internal} and {\it external} outputs. Emitting an internal
output corresponds to considering a memory,
and emitting an external output corresponds to predicting a label.
From the RNN point of view, $u$ in \fig{single}(b) and
\eqn{recurrent} is a hidden state, and the model generates an
internal output $p$ (attention weights in \fig{single}(a)) using $A$.   The model then
ingests $p$ using $C$, updates the hidden state, and so on\footnote{Note that in this view, the terminology of input and output from
 \fig{single} is flipped - when viewed as a traditional RNN with this
 special conditioning of outputs, $A$ becomes part of the output embedding of
the RNN and $C$ becomes the input embedding.}.  Here, unlike a standard RNN, we
explicitly condition on the outputs stored in memory during the $K$ hops, and we keep these
outputs soft, rather than sampling them.  Thus our model
makes several computational
steps before producing an output meant to be seen by the ``outside
world''.

% Note that if we use the layer-wise weight tying scheme, our model can
% be cast as a traditional RNN where we divide the outputs of the RNN
% into {\it internal} and {\it external} outputs.  Emitting an internal
% output corresponds to considering a memory,
% and emitting an external output corresponds to predicting a label.
% From the RNN point of view, $u$ in \fig{single}(b) and
% \eqn{recurrent} is a hidden state, and the model generates an
% internal output $p$ using $A$; here, unlike a standard RNN, we
% explicitly condition on the outputs stored in memory.  The model then
% ingests $p$ using $C$, updates the hidden state, and so on.  In
% contrast to a traditional RNN, the model makes several computational
% steps before producing an output meant to be seen by the ``outside
% world''.  In this view, the terminology of input and output from
% \fig{single} is flipped - when viewed as a traditional RNN with this
% special conditioning of outputs, $A$ becomes the output embedding of
% the RNN and $C$ becomes the input embedding.

\section{Related Work}

A number of recent efforts have explored ways to capture long-term
structure within sequences using RNNs or LSTM-based models
\cite{Chung14,Graves13,Koutnik14,Mikolov14,hochreiter1997long,Atkeson95memory-basedneural}.  The memory in
these models is the state of the network, which is latent and
inherently unstable over long timescales.  The LSTM-based models
address this through local memory cells which lock in the network
state from the past. In practice, the performance gains over carefully
trained RNNs are modest (see Mikolov \etal \cite{Mikolov14}). Our model differs from these
in that it uses a global memory, with shared read and write
functions. However, with layer-wise weight tying our model can be
viewed as a form of RNN which only produces an output after a fixed number of
time steps (corresponding to the number of hops), with the
intermediary steps involving memory input/output operations that
update the internal state.

Some of the very early work on neural networks by
Steinbuch and Piske\cite{Steinbuch63} and Taylor \cite{Taylor59} considered a memory that performed
nearest-neighbor operations on stored input vectors and then fit
parametric models to the retrieved sets. This has similarities to a
single layer version of our model.

Subsequent work in the 1990's explored other types of memory
\cite{Pollack91,Das92,mozer1993connectionist}. For example, Das \etal
\cite{Das92} and Mozer \etal \cite{mozer1993connectionist} introduced an explicit stack with
push and pop operations which has been revisited recently by \cite{Joulin15} in the context of an RNN model.

Closely related to our model is the Neural Turing Machine of
Graves \etal \cite{Graves14}, which also uses a continuous memory representation. The NTM memory uses both content and address-based access,
unlike ours which only explicitly allows the former, although the
temporal features that we will introduce in \secc{model_details_QA} allow a kind of address-based access.  However, in
part because we always write each memory sequentially, our model is
somewhat simpler, not requiring operations like
sharpening. Furthermore, we apply our memory model to textual
reasoning tasks, which qualitatively differ from the more abstract
operations of sorting and recall tackled by the NTM.

Our model is also related to Bahdanau \etal \cite{BahdanauCB14}.  In that
work, a bidirectional RNN based encoder and gated RNN based decoder were used for
machine translation. The decoder uses an attention model that finds
which hidden states from the encoding are most useful for outputting
the next translated word; the attention model uses a small neural
network that takes as input a concatenation of the current hidden state of the
decoder and each of the encoders hidden states. A similar attention model
is also used in Xu \etal \cite{Xu15} for generating image captions.
Our ``memory'' is analogous to their attention mechanism, although \cite{BahdanauCB14} is only over a single
sentence rather than many, as in our case. Furthermore, our model makes
several hops on the memory before making an output; we will see below that this is important for
good performance.  There are also
differences in the architecture of the small network used to score
the memories compared to our scoring approach; we use a simple linear layer, whereas they
use a more sophisticated gated architecture.

We will apply our model to language modeling, an extensively studied
task. Goodman \cite{Goodman01} showed simple but effective approaches which
combine $n$-grams with a cache. Bengio \etal \cite{Bengio03} ignited
interest in using neural network based models for the task, with RNNs
\cite{Mikolov12} and LSTMs \cite{hochreiter1997long,sundermeyer12}
showing clear performance gains over traditional methods. Indeed,
the current state-of-the-art is held by variants of these models,
for example very large LSTMs with Dropout \cite{Zaremba2014rnn} or
RNNs with diagonal constraints on the weight matrix \cite{Mikolov14}.  With
appropriate weight tying, our model can be regarded as a modified form
of RNN, where the recurrence is indexed by memory lookups to the word sequence rather than indexed by the sequence itself.

\section{Synthetic Question and Answering  Experiments}

We perform experiments on the synthetic QA tasks defined in
\cite{Weston15} (using version 1.1 of the dataset). A given QA task consists of a set of
statements, followed by a question whose answer is typically a single
word (in a few tasks, answers are a set of words).
The answer is available to the model at training time,
but must be predicted at test time.
There are a total of 20 different types of tasks that probe
different forms of reasoning and deduction.
Here are samples of three of the tasks:
\vspace{-1mm}
\begin{table}[h!]
\vspace{-2mm}
  \centering
  \scriptsize
\begin{tabular}{l|l|p{6cm}}
\texttt{Sam walks into the kitchen.} & \texttt{Brian is a lion.}& \texttt{Mary journeyed to the den.}\\
\texttt{Sam picks up an apple.} & \texttt{Julius is a lion.}& \texttt{Mary went back to the kitchen.}\\
\texttt{Sam walks into the bedroom.} &\texttt{Julius is white.} & \texttt{John journeyed to the bedroom.}\\
\texttt{Sam drops the apple.} & \texttt{Bernhard is green.}& \texttt{Mary discarded the milk.}\\
\texttt{\textcolor{blue}{Q: Where is the apple?}} &\texttt{\textcolor{blue}{Q: What color is Brian?}} &\texttt{\textcolor{blue}{Q: Where was the milk before the den?}} \\
\texttt{\textcolor{red}{A. Bedroom}} &\texttt{\textcolor{red}{A. White}}  &\texttt{\textcolor{red}{A. Hallway}}
\end{tabular}
\vspace{-2mm}
\label{tab:QA_examples}
\vspace{-2mm}
\end{table}

\vspace{-2mm}
Note that for each question, only some subset of the
statements contain information needed for the answer, and the
others are essentially irrelevant distractors (e.g. the first sentence in the first
example). In the Memory Networks of Weston \etal \cite{Weston15}, this {\em supporting subset} was explicitly
indicated to the model during training and the key difference between that work and this one is that this information is no longer provided. Hence, the model
must deduce for itself at training and test time which sentences are relevant and which are
not.

Formally, for one of the 20 QA tasks, we are given example
problems, each having a set of $I$ sentences $\{x_i\}$ where
$I \leq 320$; a question sentence $q$ and answer $a$. Let the $j$th word of sentence
$i$ be $x_{ij}$, represented by a one-hot vector of length $V$ (where the
vocabulary is of size $V=177$, reflecting the simplistic nature of the
QA language). The same representation is used for the question $q$
and answer $a$.
Two versions of the data are used, one that has 1000 training problems per
task and a second larger one with 10,000 per task.

\subsection{Model Details}
\label{sec:model_details_QA}

Unless otherwise stated, all experiments used a $K=3$ hops model with the adjacent weight
sharing scheme.  For all tasks that output lists
(i.e.\ the answers are multiple words), we take each possible
combination of possible outputs and record them as a separate answer
vocabulary word.

\noindent {\bf Sentence Representation:} In our experiments we explore two different representations for the
sentences. The first is the bag-of-words (BoW) representation that takes the sentence $x_i = \{x_{i1}, x_{i2},..., x_{in}\}$,
embeds each word and sums the resulting vectors: e.g~$m_i=\sum_j A x_{ij}$ and $c_i=\sum_j C x_{ij}$.   The input vector $u$ representing
the question is also embedded as a bag of words: $u = \sum_j Bq_j$.
This has the drawback that it cannot capture the order of the words
in the sentence, which is important for some tasks.

We therefore propose a second representation that encodes the position
of words within the sentence. This takes the form: $m_i = \sum_j
l_j \cdot A x_{ij}$, where $\cdot$ is an element-wise multiplication.
$l_j$ is a column vector with the structure
$l_{kj}=(1-j/J)-(k/d)(1-2j/J)$ (assuming 1-based indexing), with $J$ being the number of words in the
sentence, and $d$ is the dimension of the embedding.
This sentence representation, which we call position encoding (PE), means that the order of the words now
affects $m_i$. The same representation is used for questions, memory inputs and memory outputs.

\noindent {\bf Temporal Encoding:} Many of the QA tasks require some notion of temporal context,
i.e. in the first example of \secc{approach}, the model needs to understand
that Sam is in the bedroom after he is in the kitchen. To
enable our model to address them, we modify the memory vector so that $m_i = \sum_j Ax_{ij} + T_A(i)$, where $T_A(i)$ is the $i$th row of a
special matrix $T_A$ that encodes temporal information. The output embedding is augmented in
the same way with a matrix $T_c$ (e.g.~$c_i = \sum_j C x_{ij} + T_C(i)$). Both $T_A$ and $T_C$ are
learned during training. They are also subject to the same sharing
constraints as $A$ and $C$. Note that sentences are indexed in reverse
order, reflecting their relative distance from the question so that
$x_1$ is the last sentence of the story.

{\bf Learning time invariance by injecting random noise}:
we have found it helpful to add ``dummy'' memories to regularize $T_A$.
That is, at training time we can randomly add 10\% of empty memories to the stories.
We refer to this approach as random noise (RN).

\subsection{Training Details}
\label{sec:training}

10\% of the bAbI training set was held-out to form a validation set,
which was used to select the optimal model architecture and hyperparameters.
Our models were trained using a learning rate of $\eta=0.01$, with
anneals every 25 epochs by $\eta/2$ until 100 epochs were reached. No
momentum or weight decay was used. The weights were initialized
randomly from a Gaussian distribution with zero mean and
$\sigma=0.1$. When trained on all tasks simultaneously with
1k training samples (10k training samples), 60 epochs (20 epochs)
were used with learning rate anneals of $\eta/2$ every 15 epochs (5
epochs). All training uses a batch size of 32 (but cost is not
averaged over a batch), and gradients with an $\ell_2$ norm larger than 40
are divided by a scalar to have norm 40.   In some of our experiments, we explored commencing training with the
softmax in each memory layer removed, making the model entirely
linear except for the final softmax for answer prediction. When the validation
loss stopped decreasing, the softmax layers were re-inserted and
training recommenced. We refer to this as linear start (LS) training.
In LS training, the initial learning rate is set to $\eta= 0.005$. The capacity of memory is restricted to the most recent 50
sentences. Since the number of sentences and the number of words per
sentence varied between problems, a null symbol was used to pad them
all to a fixed size. The embedding of the null symbol was constrained
to be zero.

 On some tasks, we observed a large variance in the
performance of our model (i.e.~sometimes failing badly, other times
not, depending on the initialization). To remedy this, we repeated each training
10 times with different random initializations, and picked the one with the lowest
 training error.

\subsection{Baselines}
We compare our approach\footnote{
MemN2N source code is available at \url{https://github.com/facebook/MemNN}.} (abbreviated to MemN2N) to a range of alternate models:
\begin{itemize}
\item {\bf MemNN:}
  The strongly supervised AM+NG+NL Memory Networks
  approach, proposed in \cite{Weston15}. This is the best reported approach in that paper.
 It uses a max operation (rather than softmax) at each layer which is trained directly with supporting facts (strong supervision). It employs $n$-gram modeling, nonlinear layers and an adaptive number of hops per query.

\item {\bf MemNN-WSH:} A weakly supervised heuristic version of MemNN where the
  supporting sentence labels are not used in training. Since we are
  unable to backpropagate through the max operations in each layer,
  we enforce that the first memory hop should share at least one word with the
question, and that the second memory hop should share at least one
word with the first hop and at least one word with the answer. All those memories that
conform are called valid memories, and the goal during training is to rank them higher
than invalid memories using the same ranking criteria as during strongly
supervised training.
\item {\bf LSTM:} A standard LSTM model, trained using question /
  answer pairs only (i.e.~also weakly supervised). For more detail, see \cite{Weston15}.
\end{itemize}

\subsection{Results}
We report a variety of design choices: (i) BoW vs Position Encoding (PE)
sentence representation; (ii) training on all 20 tasks independently vs jointly
training (joint training used an embedding dimension of
$d=50$, while independent training used $d=20$); (iii) two phase
training: linear start (LS) where softmaxes are removed initially vs training with softmaxes from the start; (iv) varying memory hops from 1 to 3.

The results across all 20 tasks are given in \tab{1k} for the 1k
training set, along with the mean performance for 10k training set\footnote{More detailed results for the 10k training set can be
found in Appendix~\ref{app:babi_10k}.}.
They show a number of interesting points:
\begin{itemize}[leftmargin=0cm,itemindent=.5cm,labelwidth=\itemindent,labelsep=0cm,align=left]
\item The best MemN2N models are reasonably close to the supervised models (e.g. 1k: 6.7\% for MemNN vs 12.6\% for MemN2N with position encoding + linear start + random noise, jointly trained
and 10k: 3.2\% for MemNN vs 4.2\% for MemN2N with position encoding
+ linear start + random noise + non-linearity\footnote{Following \cite{Peng15} we found adding more non-linearity solves tasks 17 and 19, see Appendix~\ref{app:babi_10k}.},
% We learned that adding non-linearity after each hops was essential for solving task 17 and 19.}),
although the supervised models are still superior.
\vspace{-0.5mm}
\item All variants of our proposed model comfortably beat the weakly supervised baseline methods.
\vspace{-0.5mm}
\item The position encoding (PE) representation improves over bag-of-words (BoW), as demonstrated by clear improvements on tasks 4, 5, 15 and 18, where word ordering is particularly important.
\vspace{-0.5mm}
\item The linear start (LS) to training seems to help avoid local
  minima. See task 16 in \tab{1k}, where PE alone gets 53.6\% error,
  while using LS reduces it to 1.6\%.
\vspace{-0.5mm}
\item Jittering the time index with random empty memories (RN) as described in
\secc{model_details_QA}
 gives a small but consistent boost in performance, especially for the smaller 1k training set.
\vspace{-0.5mm}
\item Joint training on all tasks helps.
\vspace{-0.5mm}
\item Importantly, more computational hops give improved performance.
We give examples of the hops performed (via the values of
eq. (\ref{eq:p})) over some illustrative examples in
Fig. \ref{fig:actbaby} and in Appendix~\ref{app:babi_act}. %
\end{itemize}

\begin{table}[h!]
\vspace{-2mm}
  \centering
  \tiny
  \setlength{\tabcolsep}{5pt}
 \begin{tabular}{|l||c||c|c|c|c|c|c||c|c|c|c|c|}
 \hline
 & \multicolumn{3}{c|}{Baseline} & \multicolumn{9}{c|}{MemN2N} \\ \cline{2-13}
 & Strongly & & & &  & & PE & 1 hop & 2 hops & 3 hops & PE & PE LS \\
 & Supervised & LSTM & MemNN & &  & PE & LS & PE LS & PE LS & PE LS & LS RN & LW \\
Task & MemNN \cite{Weston15} & \cite{Weston15} & WSH & BoW & PE & LS & RN & joint & joint & joint & joint & joint \\
\hline
1: 1 supporting fact        &  0.0 & 50.0 &  0.1 &  0.6 &  0.1 &  0.2 &  0.0 &  0.8 &  0.0 &  0.1 &  0.0 &  0.1 \\
2: 2 supporting facts       &  0.0 & 80.0 & 42.8 & 17.6 & 21.6 & 12.8 &  8.3 & 62.0 & 15.6 & 14.0 & 11.4 & 18.8 \\
3: 3 supporting facts       &  0.0 & 80.0 & 76.4 & 71.0 & 64.2 & 58.8 & 40.3 & 76.9 & 31.6 & 33.1 & 21.9 & 31.7 \\
4: 2 argument relations     &  0.0 & 39.0 & 40.3 & 32.0 &  3.8 & 11.6 &  2.8 & 22.8 &  2.2 &  5.7 & 13.4 & 17.5 \\
5: 3 argument relations     &  2.0 & 30.0 & 16.3 & 18.3 & 14.1 & 15.7 & 13.1 & 11.0 & 13.4 & 14.8 & 14.4 & 12.9 \\
6: yes/no questions         &  0.0 & 52.0 & 51.0 &  8.7 &  7.9 &  8.7 &  7.6 &  7.2 &  2.3 &  3.3 &  2.8 &  2.0 \\
7: counting                 & 15.0 & 51.0 & 36.1 & 23.5 & 21.6 & 20.3 & 17.3 & 15.9 & 25.4 & 17.9 & 18.3 & 10.1 \\
8: lists/sets               &  9.0 & 55.0 & 37.8 & 11.4 & 12.6 & 12.7 & 10.0 & 13.2 & 11.7 & 10.1 &  9.3 &  6.1 \\
9: simple negation          &  0.0 & 36.0 & 35.9 & 21.1 & 23.3 & 17.0 & 13.2 &  5.1 &  2.0 &  3.1 &  1.9 &  1.5 \\
10: indefinite knowledge    &  2.0 & 56.0 & 68.7 & 22.8 & 17.4 & 18.6 & 15.1 & 10.6 &  5.0 &  6.6 &  6.5 &  2.6 \\
11: basic coreference       &  0.0 & 38.0 & 30.0 &  4.1 &  4.3 &  0.0 &  0.9 &  8.4 &  1.2 &  0.9 &  0.3 &  3.3 \\
12: conjunction             &  0.0 & 26.0 & 10.1 &  0.3 &  0.3 &  0.1 &  0.2 &  0.4 &  0.0 &  0.3 &  0.1 &  0.0 \\
13: compound coreference    &  0.0 &  6.0 & 19.7 & 10.5 &  9.9 &  0.3 &  0.4 &  6.3 &  0.2 &  1.4 &  0.2 &  0.5 \\
14: time reasoning          &  1.0 & 73.0 & 18.3 &  1.3 &  1.8 &  2.0 &  1.7 & 36.9 &  8.1 &  8.2 &  6.9 &  2.0 \\
15: basic deduction         &  0.0 & 79.0 & 64.8 & 24.3 &  0.0 &  0.0 &  0.0 & 46.4 &  0.5 &  0.0 &  0.0 &  1.8 \\
16: basic induction         &  0.0 & 77.0 & 50.5 & 52.0 & 52.1 &  1.6 &  1.3 & 47.4 & 51.3 &  3.5 &  2.7 & 51.0 \\
17: positional reasoning    & 35.0 & 49.0 & 50.9 & 45.4 & 50.1 & 49.0 & 51.0 & 44.4 & 41.2 & 44.5 & 40.4 & 42.6 \\
18: size reasoning          &  5.0 & 48.0 & 51.3 & 48.1 & 13.6 & 10.1 & 11.1 &  9.6 & 10.3 &  9.2 &  9.4 &  9.2 \\
19: path finding            & 64.0 & 92.0 &100.0 & 89.7 & 87.4 & 85.6 & 82.8 & 90.7 & 89.9 & 90.2 & 88.0 & 90.6 \\
20: agent's motivation      &  0.0 &  9.0 &  3.6 &  0.1 &  0.0 &  0.0 &  0.0 &  0.0 &  0.1 &  0.0 &  0.0 &  0.2 \\ \hline
Mean error (\%)             &  6.7 & 51.3 & 40.2 & 25.1 & 20.3 & 16.3 & 13.9 & 25.8 & 15.6 & 13.3 & 12.4 & 15.2 \\
Failed tasks (err. $> 5\%$) &    4 &   20 &   18 &   15 &   13 &   12 &   11 &   17 &   11 &   11 &   11 &   10 \\ \hline \hline
On 10k training data        & & & & & & & & & & & & \\
Mean error (\%)             &  3.2 & 36.4 & 39.2 & 15.4 &  9.4 &  7.2 &  6.6 & 24.5 & 10.9 &  7.9 &  7.5 & 11.0 \\
Failed tasks (err. $> 5\%$) &    2 &   16 &   17 &    9 &    6 &    4 &    4 &   16 &    7 &    6 &    6 &    6 \\
\hline
\end{tabular}

\vspace{-1mm}
  \caption{Test error rates (\%) on the 20 QA tasks for models using
    1k training examples (mean test errors for 10k training examples are shown at the bottom). Key: BoW = bag-of-words representation; PE =
  position encoding representation; LS = linear start training; RN = random injection of time
index noise;  LW = RNN-style layer-wise weight tying (if not stated, adjacent weight tying is used); joint = joint
  training on all tasks (as opposed to per-task training). }
\label{tab:1k}
\end{table}

\begin{figure}[h!]
\begin{center}
\includegraphics[width=1\linewidth,trim=12mm 217mm 12mm 12mm,clip=true]{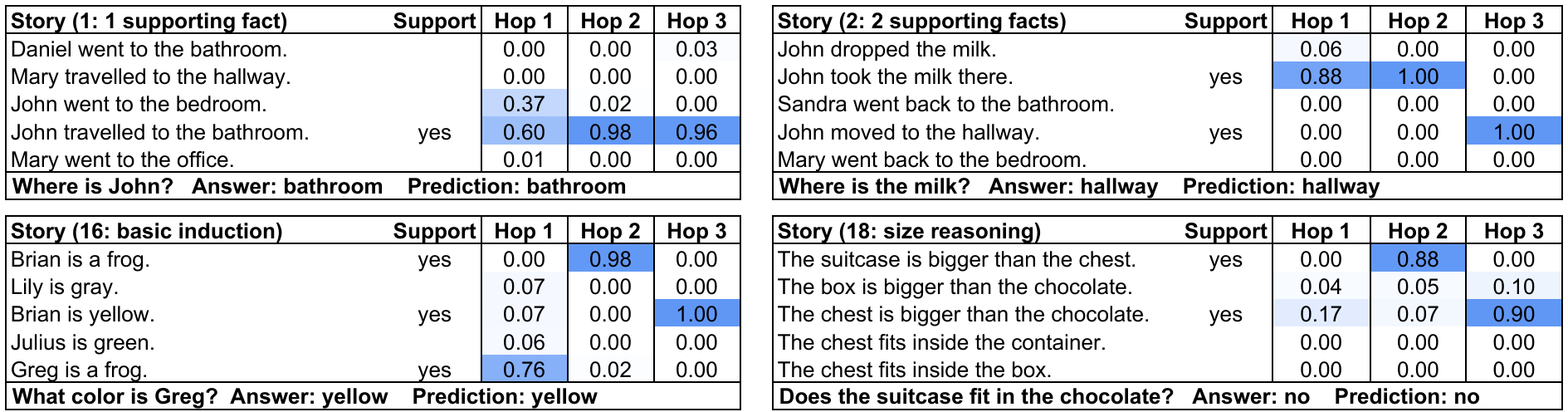}
\end{center}
\caption{Example predictions on the QA tasks of \cite{Weston15}. We show the labeled supporting facts (support) from the dataset which MemN2N does not use during training,
and the probabilities $p$ of each hop used by the model during inference.
MemN2N successfully learns to focus on the correct supporting sentences.
\label{fig:actbaby}
}
\end{figure}

\begin{table}[h!]
\small
\centering
\begin{tabular}{|l|ccccc|ccccc|}
\hline
 & \multicolumn{5}{c|}{Penn Treebank} & \multicolumn{5}{c|}{Text8} \\
& \# of & \# of & memory & Valid. & Test & \# of & \# of & memory & Valid. & Test \\
Model & hidden & hops & size & perp. & perp. & hidden & hops & size & perp. & perp. \\ \hline \hline
RNN  \cite{Mikolov14} & 300 & - & - & 133 & 129 & 500 & - & - &   - & 184 \\
LSTM \cite{Mikolov14} & 100 & - & - & 120 & 115 & 500 & - & - & 122 & 154 \\
SCRN \cite{Mikolov14} & 100 & - & - & 120 & 115 & 500 & - & - &   - & 161 \\
\hline \hline
MemN2N
 & 150 & 2 & 100 & 128 & 121 & 500 & 2 & 100 & 152 & 187 \\
 & 150 & 3 & 100 & 129 & 122 & 500 & 3 & 100 & 142 & 178 \\
 & 150 & 4 & 100 & 127 & 120 & 500 & 4 & 100 & 129 & 162 \\
 & 150 & 5 & 100 & 127 & 118 & 500 & 5 & 100 & 123 & 154 \\
 & 150 & 6 & 100 & 122 & 115 & 500 & 6 & 100 & 124 & 155 \\
 & 150 & 7 & 100 & 120 & 114 & 500 & 7 & 100 & 118 & \textbf{147} \\ \cline{2-11}
 & 150 & 6 &  25 & 125 & 118 & 500 & 6 &  25 & 131 & 163 \\
 & 150 & 6 &  50 & 121 & 114 & 500 & 6 &  50 & 132 & 166 \\
 & 150 & 6 &  75 & 122 & 114 & 500 & 6 &  75 & 126 & 158 \\
 & 150 & 6 & 100 & 122 & 115 & 500 & 6 & 100 & 124 & 155 \\
 & 150 & 6 & 125 & 120 & 112 & 500 & 6 & 125 & 125 & 157 \\
 & 150 & 6 & 150 & 121 & 114 & 500 & 6 & 150 & 123 & 154 \\ \cline{2-11}
 & 150 & 7 & 200 & 118 & \textbf{111} & - & - & - & - & - \\
\hline
\end{tabular}
\caption{The perplexity on the test sets of Penn Treebank and Text8 corpora.  Note that increasing the number of memory hops improves
performance.}
\label{tab:lang}
\end{table}

\begin{figure}[h!]
\vspace{-3mm}
\begin{center}
\includegraphics[scale=0.7,trim=0mm 0mm 20mm 0mm,clip=true]{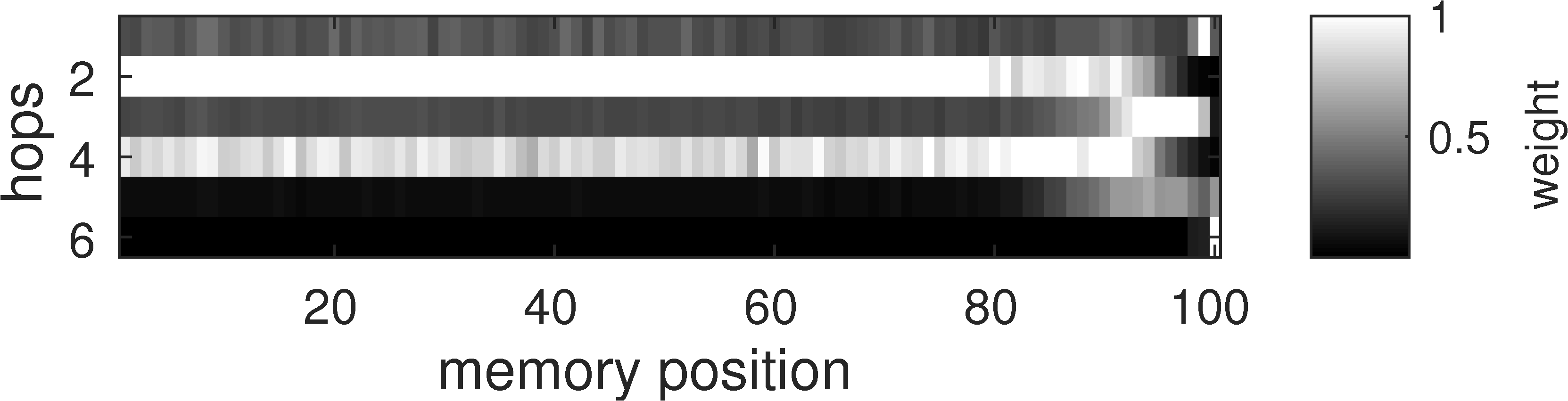}
\includegraphics[scale=0.7,trim=7mm 0mm 0mm 0mm,clip=true]{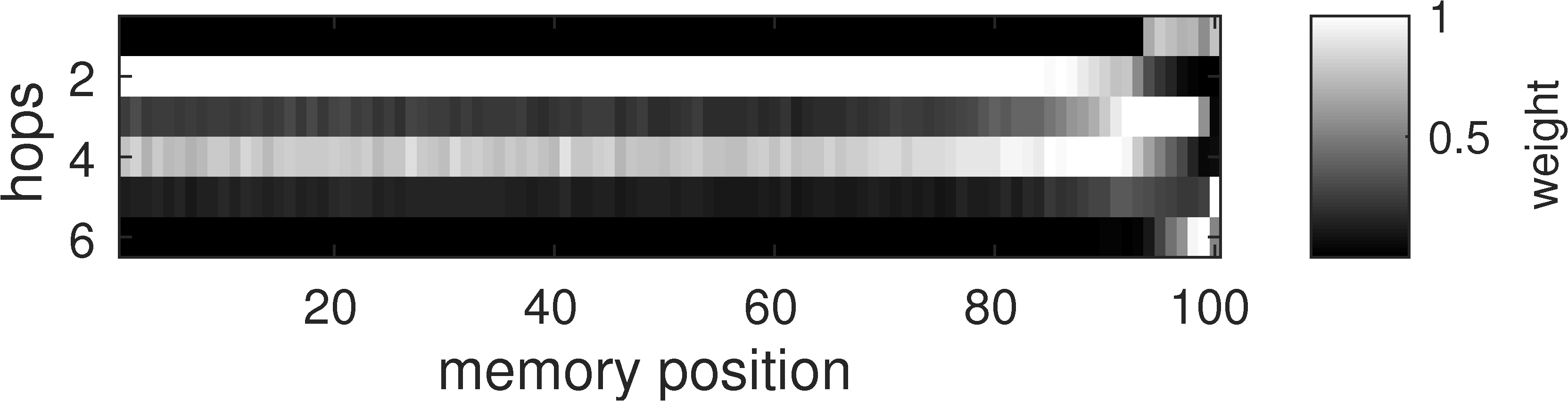}
\end{center}
\vspace*{-0.3cm}
\caption{Average activation weight of memory positions during 6 memory hops.
White color indicates where the model is attending during the $k^{th}$ hop.
For clarity, each row is normalized to have maximum value of 1.
A model is trained on (left) Penn Treebank and (right) Text8 dataset.}
\label{fig:avgact}
\vspace{-3mm}
\end{figure}

\section{Language Modeling Experiments}
The goal in language modeling is to predict the next word in a
text sequence given the previous words $x$. We now explain how our model
can easily be applied to this task.

We now operate on word level, as opposed to
the sentence level. Thus the previous $N$ words in the sequence (including the current) are
embedded into memory separately.  Each memory
cell holds only a single word, so there is no need for the BoW or
linear mapping representations used in the QA tasks. We employ the temporal embedding approach
of \secc{model_details_QA}.

Since there is
no longer any question, $q$ in \fig{single} is fixed to a constant vector 0.1 (without embedding).
The output softmax predicts which word in the vocabulary (of size $V$)
is next in the sequence. A cross-entropy loss is used to train model
by backpropagating the error through multiple memory layers, in the
same manner as the QA tasks. To aid training, we apply ReLU
operations to half of the units in each layer.
We use layer-wise (RNN-like) weight sharing, i.e.~the query weights of
each layer are the same; the output weights of each layer are the
same. As noted in \secc{multi}, this makes our architecture closely
related to an RNN which is traditionally used for language modeling
tasks; however here the ``sequence'' over which the network is recurrent is not in the text,
but in the memory hops. Furthermore,
the weight tying restricts the number of parameters in the model,
helping generalization for the deeper models which we find to be
effective for this task.
We use two different datasets:

\vspace{-1mm}
\noindent {\bf Penn Tree Bank} \cite{Marcus93}: This consists of
929k/73k/82k train/validation/test words, distributed over a
vocabulary of 10k words. The same preprocessing as
\cite{Zaremba2014rnn} was used.

\vspace{-1mm}
\noindent {\bf Text8} \cite{Mikolov14}: This is a a
pre-processed version of the first 100M million characters, dumped
from Wikipedia. This is split into 93.3M/5.7M/1M character train/validation/test
sets. All word occurring less than 5 times are replaced with the $<$UNK$>$
token, resulting in a vocabulary size of $\sim$44k.

\vspace{-1mm}
\subsection{Training Details}
\vspace{-2mm}
 The training procedure we use is the same as the QA tasks, except for the following.
For each mini-batch update, the $\ell_2$ norm of
the whole gradient of all parameters is measured\footnote{In the QA tasks, the
gradient of each weight matrix
    is measured separately.} and if larger than $L=50$, then it is scaled down to
  have norm $L$. This was crucial for good performance. We use the learning rate annealing schedule from \cite{Mikolov14}, namely, if the validation cost has not decreased after one
  epoch, then the learning rate is scaled down by a factor
  1.5. Training terminates when the learning rate drops below
  $10^{-5}$, i.e.~after 50 epochs or so.
Weights are initialized using $\mathcal{N}(0,0.05)$ and batch size is set to 128.
On the Penn tree dataset, we repeat each training 10 times with different random initializations
and pick the one with smallest validation cost. However, we have done only a single training run on Text8 dataset due to limited time constraints.

\vspace{-1mm}
\subsection{Results}
\vspace{-2mm}
\tab{lang} compares our model to RNN, LSTM and Structurally
Constrained Recurrent Nets (SCRN) \cite{Mikolov14} baselines on the two
benchmark datasets. Note that the baseline architectures were tuned in
\cite{Mikolov14} to give optimal perplexity\footnote{They tuned the hyper-parameters on Penn Treebank and used them on Text8 without additional tuning, except for the number of hidden units. See \cite{Mikolov14} for more detail.}.
Our MemN2N approach achieves lower perplexity on  both datasets (111 vs 115 for RNN/SCRN on Penn and 147 vs 154
for LSTM on Text8).
Note that MemN2N has $\sim$1.5x more parameters than RNNs with the same number of hidden units,
while LSTM has $\sim$4x more parameters.
We also vary the number of hops and memory size of our MemN2N, showing
the contribution of both to performance; note in particular that increasing the number of hops helps. In \fig{avgact}, we show how MemN2N operates
on memory with multiple hops. It shows the average weight of the activation of each memory position over the test set. We can see that some hops
concentrate only on recent words, while other hops have more broad attention over all memory locations,
which is consistent with the idea that succesful language models consist of a smoothed $n$-gram model and a cache~\cite{Mikolov14}.
Interestingly, it seems that those two types of hops tend to alternate.  Also note that unlike a traditional RNN,
the cache does not decay exponentially:  it has roughly the same average activation across the entire memory.  This may
be the source of the observed improvement in language modeling.

\vspace{-2mm}
\section{Conclusions and Future Work}
\vspace{-2mm}
 In this work we showed that a neural network with an explicit memory
and a recurrent attention mechanism for reading the memory can be
successfully trained via backpropagation on diverse tasks from
question answering to language modeling.  Compared to the Memory
Network implementation of~\cite{Weston14} there is no supervision of
supporting facts and so our model can be used in a wider range of
settings.  Our model approaches the same performance of that model,
and is significantly better than other baselines with the same level
of supervision.  On language modeling tasks, it slightly outperforms
tuned RNNs and LSTMs of comparable complexity.  On both tasks we can
see that increasing the number of memory hops improves performance.

However, there is still much to do.  Our model is still unable to
exactly match the performance of the memory networks trained with
strong supervision, and both fail on several of the 1k QA tasks.
Furthermore, smooth lookups may not scale well to the case where a
larger memory is required.  For these settings, we plan to explore
multiscale notions of attention or hashing, as proposed in
\cite{Weston14}.

\vspace{-2mm}
\section*{Acknowledgments}
\vspace{-2mm}
The authors would like to thank Armand Joulin, Tomas Mikolov, Antoine
Bordes and Sumit Chopra for useful comments and valuable
discussions, and also the FAIR Infrastructure
team for their help and support.

\bibliography{bibliography}
\bibliographystyle{ieee}

\begin{appendices}
\newpage
\section{Results on 10k QA dataset}
\label{app:babi_10k}
\begin{table}[h!]
\vspace{-2mm}
  \centering
  \tiny
  \setlength{\tabcolsep}{4pt}
 \begin{tabular}{|l||c||c|c|c|c|c|c|c||c|c|c|c|c|}
 \hline
 & \multicolumn{3}{c|}{Baseline} & \multicolumn{10}{c|}{MemN2N} \\ \cline{2-14}
 & Strongly & & & &  & & PE & PE LS & 1 hop & 2 hops & 3 hops & PE & PE LS \\
 & Supervised &  & MemNN & &  & PE & LS & LW & PE LS & PE LS & PE LS & LS RN & LW \\
Task & MemNN & LSTM & WSH & BoW & PE & LS & RN & RN$^\ast$ & joint & joint & joint & joint & joint \\
\hline
1: 1 supporting fact        &  0.0 &  0.0 &  0.1 &  0.0 &  0.0 &  0.0 &  0.0 &  0.0 &  0.0 &  0.0 &  0.0 &  0.0 &  0.0 \\
2: 2 supporting facts       &  0.0 & 81.9 & 39.6 &  0.6 &  0.4 &  0.5 &  0.3 &  0.3 & 62.0 &  1.3 &  2.3 &  1.0 &  0.8 \\
3: 3 supporting facts       &  0.0 & 83.1 & 79.5 & 17.8 & 12.6 & 15.0 &  9.3 &  2.1 & 80.0 & 15.8 & 14.0 &  6.8 & 18.3 \\
4: 2 argument relations     &  0.0 &  0.2 & 36.6 & 31.8 &  0.0 &  0.0 &  0.0 &  0.0 & 21.4 &  0.0 &  0.0 &  0.0 &  0.0 \\
5: 3 argument relations     &  0.3 &  1.2 & 21.1 & 14.2 &  0.8 &  0.6 &  0.8 &  0.8 &  8.7 &  7.2 &  7.5 &  6.1 &  0.8 \\
6: yes/no questions         &  0.0 & 51.8 & 49.9 &  0.1 &  0.2 &  0.1 &  0.0 &  0.1 &  6.1 &  0.7 &  0.2 &  0.1 &  0.1 \\
7: counting                 &  3.3 & 24.9 & 35.1 & 10.7 &  5.7 &  3.2 &  3.7 &  2.0 & 14.8 & 10.5 &  6.1 &  6.6 &  8.4 \\
8: lists/sets               &  1.0 & 34.1 & 42.7 &  1.4 &  2.4 &  2.2 &  0.8 &  0.9 &  8.9 &  4.7 &  4.0 &  2.7 &  1.4 \\
9: simple negation          &  0.0 & 20.2 & 36.4 &  1.8 &  1.3 &  2.0 &  0.8 &  0.3 &  3.7 &  0.4 &  0.0 &  0.0 &  0.2 \\
10: indefinite knowledge    &  0.0 & 30.1 & 76.0 &  1.9 &  1.7 &  3.3 &  2.4 &  0.0 & 10.3 &  0.6 &  0.4 &  0.5 &  0.0 \\
11: basic coreference       &  0.0 & 10.3 & 25.3 &  0.0 &  0.0 &  0.0 &  0.0 &  0.1 &  8.3 &  0.0 &  0.0 &  0.0 &  0.4 \\
12: conjunction             &  0.0 & 23.4 &  0.0 &  0.0 &  0.0 &  0.0 &  0.0 &  0.0 &  0.0 &  0.0 &  0.0 &  0.1 &  0.0 \\
13: compound coreference    &  0.0 &  6.1 & 12.3 &  0.0 &  0.1 &  0.0 &  0.0 &  0.0 &  5.6 &  0.0 &  0.0 &  0.0 &  0.0 \\
14: time reasoning          &  0.0 & 81.0 &  8.7 &  0.0 &  0.2 &  0.0 &  0.0 &  0.1 & 30.9 &  0.2 &  0.2 &  0.0 &  1.7 \\
15: basic deduction         &  0.0 & 78.7 & 68.8 & 12.5 &  0.0 &  0.0 &  0.0 &  0.0 & 42.6 &  0.0 &  0.0 &  0.2 &  0.0 \\
16: basic induction         &  0.0 & 51.9 & 50.9 & 50.9 & 48.6 &  0.1 &  0.4 & 51.8 & 47.3 & 46.4 &  0.4 &  0.2 & 49.2 \\
17: positional reasoning    & 24.6 & 50.1 & 51.1 & 47.4 & 40.3 & 41.1 & 40.7 & 18,6 & 40.0 & 39.7 & 41.7 & 41.8 & 40.0 \\
18: size reasoning          &  2.1 &  6.8 & 45.8 & 41.3 &  7.4 &  8.6 &  6.7 &  5.3 &  9.2 & 10.1 &  8.6 &  8.0 &  8.4 \\
19: path finding            & 31.9 & 90.3 &100.0 & 75.4 & 66.6 & 66.7 & 66.5 &  2.3 & 91.0 & 80.8 & 73.3 & 75.7 & 89.5 \\
20: agent's motivation      &  0.0 &  2.1 &  4.1 &  0.0 &  0.0 &  0.0 &  0.0 &  0.0 & 0.0 &  0.0 &  0.0 &  0.0 &  0.0 \\ \hline
Mean error (\%)             &  3.2 & 36.4 & 39.2 & 15.4 &  9.4 &  7.2 &  6.6 &  4.2 & 24.5 & 10.9 &  7.9 &  7.5 & 11.0 \\
Failed tasks (err. $> 5\%$) &    2 &   16 &   17 &    9 &    6 &    4 &    4 &    3 &   16 &    7 &    6 &    6 &    6 \\
\hline
\end{tabular}

\vspace{-1mm}
  \caption{Test error rates (\%) on the 20 bAbI QA tasks for models using
    10k training examples. Key: BoW = bag-of-words representation; PE =
  position encoding representation; LS = linear start training; RN = random injection of time
index noise;  LW = RNN-style layer-wise weight tying (if not stated, adjacent weight tying is used); joint = joint
  training on all tasks (as opposed to per-task training); $\ast$ = this is a larger model with non-linearity (embedding dimension is $d=100$ and ReLU applied to the internal state after each hop. This was inspired by \cite{Peng15} and crucial for getting better performance on tasks 17 and 19). }
\label{tab:10k}
\end{table}

\newpage
\section{Visualization of attention weights in QA problems}
\label{app:babi_act}

\begin{figure}[h!]
\vspace{-3mm}
\begin{center}
\includegraphics[width=1\linewidth,trim=12mm 13mm 12mm 12mm,clip=true]{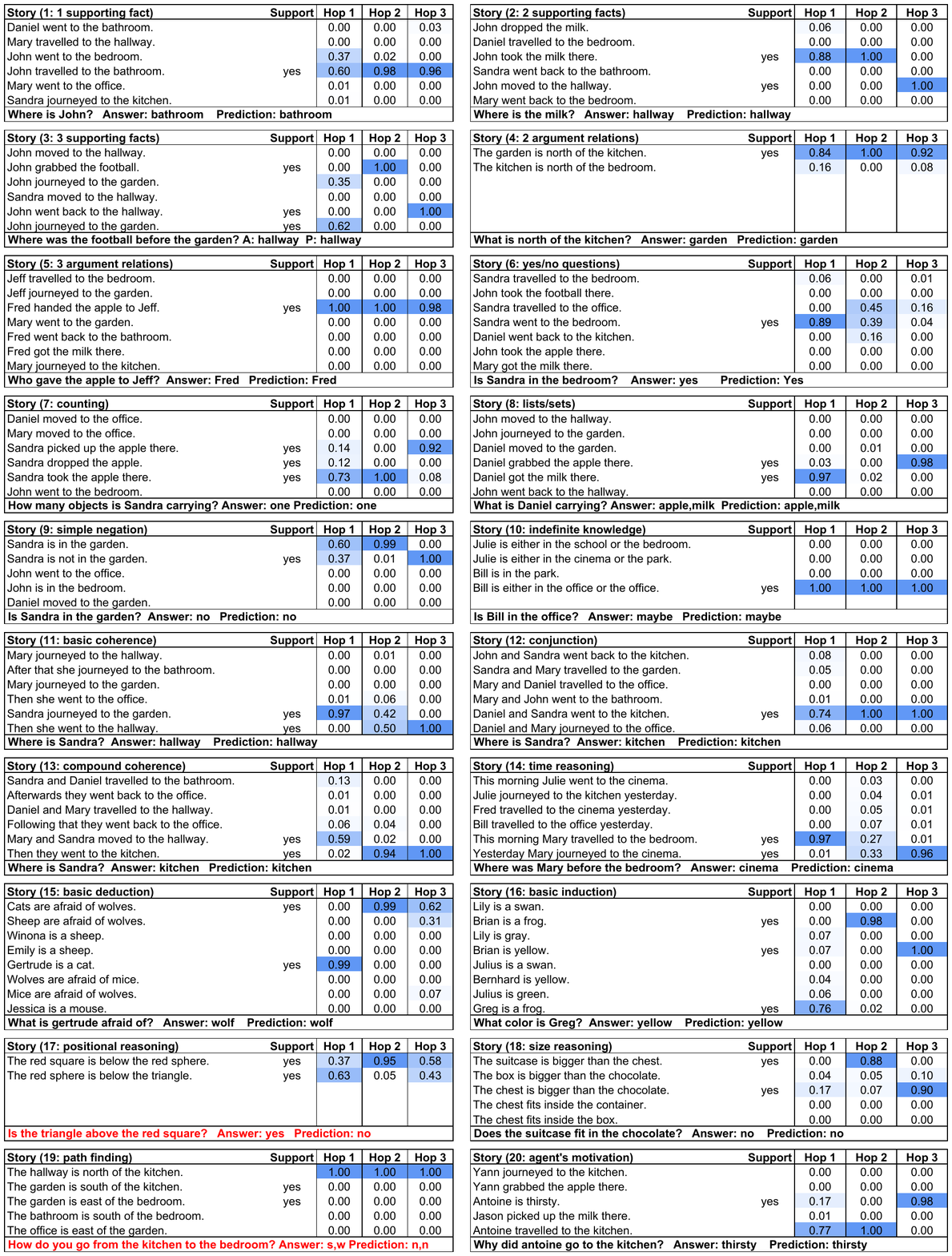}
\end{center}
\vspace*{-0.3cm}
\caption{Examples of attention weights during different memory hops for the bAbi tasks.
The model is PE+LS+RN with 3 memory hops that is trained separately on each task with 10k training data.
The support column shows which sentences are necessary for answering questions.
Although this information is not used, the model succesfully learns to focus on the correct support sentences
on most of the tasks. The hop columns show where the model put more weight (indicated by values and blue color)
during its three hops. The mistakes made by the model are highlighted by red color.}
\label{fig:actbabi}
\vspace{-3mm}
\end{figure}

\end{appendices}

\end{document}